%% file: paper.tex
\documentclass[sigconf]{acmart}

\usepackage{hyperref}
\usepackage{multirow}
\usepackage{enumitem}
\usepackage[T1]{fontenc}

\usepackage{mathabx}
\def\CircleArrowright{\ensuremath{
  \rotatebox[origin=c]{310}{$\circlearrowright$}}}
\newcommand{\rvlnbert}{VLN$\protect\CircleArrowright$BERT}
\usepackage{pifont}
\newcommand{\high}[1]{{\textbf{\color[RGB]{0,0,255}#1}}}

\AtBeginDocument{}
\copyrightyear{2023}
\acmYear{2023}
\setcopyright{none}
\acmConference[MM '23] {Proceedings of the 31st ACM International Conference on Multimedia}{October 29--November 3, 2023}{Ottawa, ON, Canada.}
\acmBooktitle{Proceedings of the 31st ACM International Conference on Multimedia (MM '23), October 29--November 3, 2023, Ottawa, ON, Canada}
\renewcommand\footnotetextcopyrightpermission[1]{} 

\begin{document}

\title{Mind the Gap: Improving Success Rate of Vision-and-Language Navigation by Revisiting Oracle Success Routes}

\settopmatter{authorsperrow=2}
\author{Chongyang Zhao}
\affiliation{
\department{Australian Institute for Machine Learning}
\institution{The University of Adelaide}
\city{Adelaide}
\country{Australia}
}
\email{chongyang.zhao@adeaide.edu.au}

\author{Yuankai Qi}
\affiliation{
\department{Australian Institute for Machine Learning}
\institution{The University of Adelaide}
\city{Adelaide}
\country{Australia}
}
\email{qykshr@gmail.com}

\author{Qi Wu}
\authornote{Corresponding author.}
\affiliation{
\department{Australian Institute for Machine Learning}
\institution{The University of Adelaide}
\city{Adelaide}
\country{Australia}
}
\email{qi.wu01@adeaide.edu.au}

\begin{abstract}
Vision-and-Language Navigation (VLN) aims to navigate to the target location by following a given instruction. Unlike existing methods focused on predicting a more accurate action at each step in navigation, in this paper, we make the first attempt to tackle a long-ignored problem in VLN: narrowing the gap between Success Rate (SR) and Oracle Success Rate (OSR). We observe a consistently large gap (up to 9\%) on four state-of-the-art VLN methods across two benchmark datasets: R2R and REVERIE. The high OSR indicates the robot agent passes the target location, while the low SR suggests the agent actually fails to stop at the target location at last. Instead of predicting actions directly, we propose to mine the target location from a trajectory given by off-the-shelf VLN models. Specially, we design a multi-module transformer-based model for learning compact discriminative trajectory viewpoint representation, which is used to predict the confidence of being a target location as described in the instruction. The proposed method is evaluated on three widely-adopted datasets: R2R, REVERIE and NDH, and shows promising results, demonstrating the potential for more future research.
\end{abstract}




\maketitle

\section{Introduction}
\begin{figure}[!t]
\centering 
\includegraphics[width=0.88\linewidth]{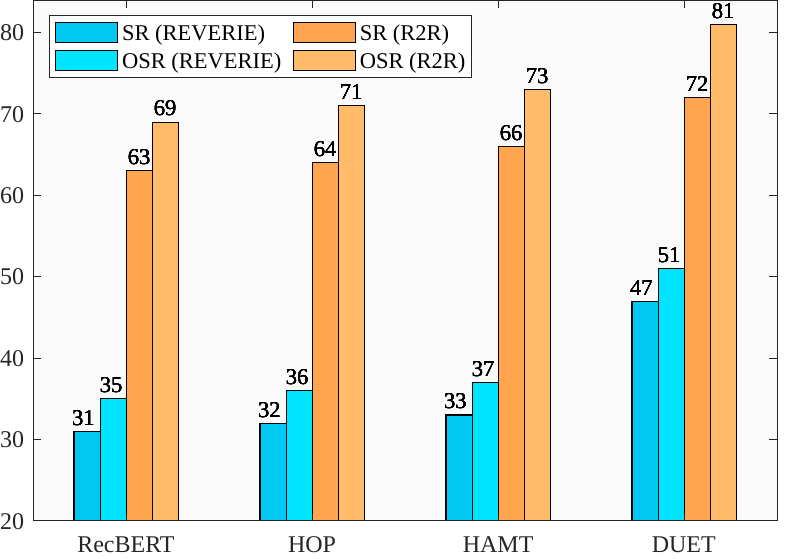}
\caption{Significant gap exists between success rate (SR) and oracle success rate (OSR) of state-of-the-art methods across VLN tasks, Room-to-Room\cite{r2r} and REVERIE \cite{reverie} on validation unseen splits.
}
\label{fig:gap}
\end{figure}
Vision-and-Language Navigation (VLN) has received increasing attention in computer vision, natural language processing, and robotics communities due to its great potential for real-world applications such as domestic assistants. The VLN task requires an agent to navigate to a target location in a 3D simulated environment based on visual observation and a given natural language instruction. A variety of VLN tasks have been proposed, including indoor room-to-room navigation according to detailed instructions (\textit{e.g.}, R2R~\cite{r2r} and RxR~\cite{rxr}) or dialogue-based instructions (\textit{e.g.}, NDH~\cite{ndh}); remote object navigation according to concise instructions (REVERIE~\cite{reverie}) or detailed instructions (SOON~\cite{soon}). Currently, there are both discrete~\cite{r2r} and continuous~\cite{vlnce, vxn} simulators. Our work is based on the discrete simulator, which provides predefined navigation graphs, thus encouraging  researchers to focus on textual-visual alignment, excluding the impact from tasks like topology mapping and obstacle avoidance. 

Most existing methods formulate VLN as a sequential text-image matching problem. To be specific, positioned at a node of a navigation graph, the agent traverses the environment by selecting one of its neighbouring nodes (represented by images) that has the maximal similarity to instruction. To improve the performance of visual-textual matching, many approaches have been developed. Data augmentation is explored to overcome the scarcity of training data via generating pseudo trajectory-instruction pairs~\cite{speakerfollower,prompt}, collecting extra data~\cite{airbert,hm3d}, and editing environments from different houses~\cite{envedit,envmixup}.  In~\cite{nvem,disentangle22,graph,oaam}, instructions are decomposed into landmark, action, and scene components via the attention mechanism to enable fine-grained cross-modality matching. Vision-language pretraining has been employed in previous works such as~\cite{prevalent, recurrent, airbert, orist, hamt, hop, hop+, duet}. Most recently, memory and maps for navigation history and planning have been utilized to further enhance textual-visual matching in the works of~\cite{hamt, duet, varmem22, hop+,gao2023adaptive,zhao2022target, bevbert,wang2023lana}.

Although these methods have boosted the VLN performance, there is one problem that remains open and has long been ignored: the performance gap between Success Rate (SR, the agent stops at a graph node within 3 meters to the target location) and Oracle Success Rate (OSR, the agent passes by or stops at a graph node within 3 meters to the destination). Instead of only happening to one specific method, this gap exists among most existing state-of-the-art methods and across datasets. Figure~\ref{fig:gap} presents the statistics of SR and OSR of four cutting-edge methods on two popular VLN tasks R2R and REVERIE. It shows that the performance gap ranges from 2.9\% to 4.4\% on the REVERIE task and consistently remains around 7-9\% for all four methods on the R2R task. This significant gap suggests existing VLN models may successfully find the correct trajectory (because it passes the target location.), however fail to stop at the target spot finally. Such a large gap demonstrates a great potential to advance VLN performance, and why not mine the passed target nodes in trajectories?

In this paper, we study how to find the target node in a given trajectory. Interestingly, we find this task could be formulated as a video grounding~\cite{tall, zhou2018towards, krishna2017dense, vidstg, hcstvg} task, \textit{i.e.}, localise the target frame (location) from a given video (trajectory) based on a given textual description (instruction). However, there are still two significant differences: (a) each node in a trajectory is represented by a panorama (usually consists of 3$\times$12 discrete observation views, \textit{i.e.}, 3 elevation levels and 12 views at each elevation) and panoramas between two consecutive nodes present sever instant visual transition (\textit{i.e.}, it is full of cutaway). (b) There may be only a few images (usually less than three) that match the textual description for the destination, which is much less than that in the video grounding.

To overcome these challenges, we design a transformer-based model consisting of three modules: Cross-Modality Elevation Transformer, Spatial-Temporal Transformer, and Target Selection Transformer. The Cross-Modality Elevation Transformer learns to fuse visual perceptions across three elevations, which helps reduce redundant information. The Spatial-Temporal Transformer is used to exchange information across different time steps and heading views at each step. The Target Selection Transformer takes advantage of learnable queries to summarize visual information at each time step and generate the representation for confidence prediction of being the target location. In summary, our contributions are three-fold: 
\begin{itemize}[leftmargin=16pt]
\item To the best of our knowledge, this is the first work to study the SR and OSR gap issue on the VLN tasks.
\item We design a multi-module transformer-based model to minimize the SR and OSR gap.
\item Extensive experiments on three popular VLN benchmarks: R2R, REVERIE and NDH, show the effectiveness of the proposed method, indicating the possibility for more future research.
\end{itemize}

\begin{figure*}[!t]
\centering
\includegraphics[width=0.98\linewidth]{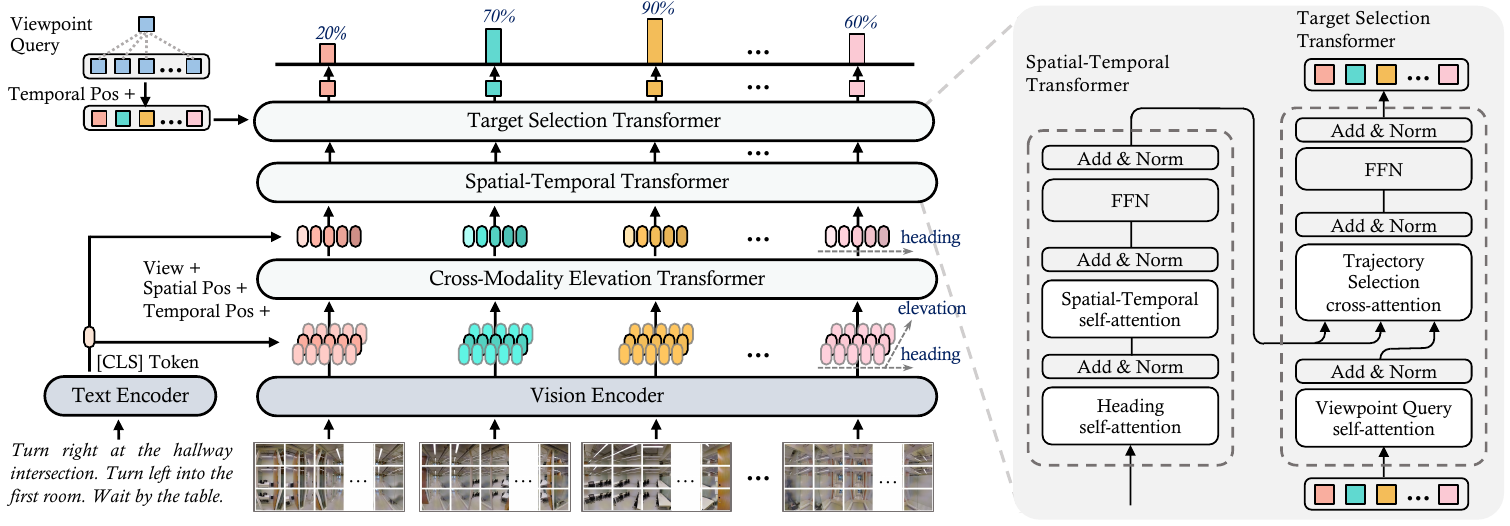}
\caption{The main architecture of the proposed trajectory grounding model. The model takes an instruction-trajectory pair as input and utilizes a text encoder and a vision encoder to extract their own representations $\langle {\mathbf {t}_{cls}, \mathcal{O}'_1}, \cdots, {\mathcal{O}'_T} \rangle$. The Cross-Modality Elevation Transformer module integrates information across different elevations while taking into account the contextual guidance provided by the instruction description. Then, the Spatial-Temporal Transformer exchanges information across discrete views and temporal navigation steps. Finally, the Target Selection Transformer summarizes the information of each viewpoint using learnable queries and predicts the confidence of each viewpoint as the destination.}
\label{fig:architecture}
\end{figure*}

\section{Related Work}
\noindent\textbf{Vision-and-Language Navigation.}
The R2R~\cite{r2r}, REVERIE~\cite{reverie}, and NDH~\cite{ndh} are three well-known benchmarks in the field of VLN, each presenting distinctive challenges for navigation agents. R2R offers low-level instructions in natural language and photo-realistic environments for navigation purposes. REVERIE presents the task of remote object localization using concise high-level instructions. NDH focuses on dialog-based navigation. A number of existing works have focused on visual-textual matching to tackle the VLN tasks. Ma \textit{et al.}~\cite{selfmonitor} propose a visual-textual co-grounding module and a progress monitor to estimate progress towards the goal. Qi \textit{et al.}~\cite{oaam} use an object- and action-aware module framework to match the decomposed instruction and candidate visual features. Hong \textit{et al.}~\cite{graph} model inter-and-intra relationships among the scene, objects and directions via a language and visual entity relation graph. Recent works in VLN have leveraged transformer-based architectures to encode vision and language information. PRESS~\cite{press} adopts BERT ~\cite{bert}, while~\cite{recurrent} reuses the build-in $\texttt{[CLS]}$ token to maintain history information recurrently. To improve generalization ability, data augmentation has been applied in~\cite{speakerfollower, envdrop, envmixup, envedit,wang2023scaling,li2023panogen}. Furthermore, adversarial learning was utilized to mine hard training samples in~\cite{advsampling}, while~\cite{airbert, hm3d} leverage additional training data to improve performance. ADAPT~\cite{adapt} proposes modality-aligned action prompts to effectively guide agents in completing complex navigation tasks. CCC~\cite{ccc} explores the intrinsic correlation between instruction following and instruction generation by jointly learning both tasks. CITL~\cite{citl} employs contrastive learning to acquire the alignment between trajectory and instruction. SIG~\cite{sig} generates potential future views to benefit the agent during navigation. History memories are specially maintained in~\cite{hamt, duet, hop+,wang2021structured, navgpt} to capture long-range dependencies across past observations and actions. All the above methods have advanced the progress of VLN. However, they all suffer from a performance gap between success rate and oracle success rate, indicating they pass by the target location but fail to stop on it. In this paper, we design a multi-module transformer-based model to mitigate this issue and hope to give hints for future work.

~\newline\noindent\textbf{Video Grounding.}
Visual grounding involves localizing an object using a referring expression, researched extensively in both image domain~\cite{qiao2021rec,mattnet,deng2022visual,transvg} and video domain. For video grounding, two tasks are distinguished: Spatial-Temporal Video Grounding (STVG) aims to localize both the frames and bounding boxes of an object specified by a language description~\cite{yamaguchi2017spatio,stvgbert,tubedetr}, and Temporal Video Grounding (TVG) that only localizes frames~\cite{chen2018,rodriguez2020}. Our formulation of VLN, localizing the image representing the navigation destination in an image sequence of the navigation trajectory, can be reformulated as part of the TVG task. Early TVG approaches employ a proposal-based architecture~\cite{tall,chen2018,yuan2019,multilevel,man}, and several proposal-free methods~\cite{yuan2019aaai,rodriguez2020,mun2020} are proposed to reduce the computation cost of proposal feature exacting. Techniques such as attention mechanism~\cite{rodriguez2020, yuan2019aaai,mun2020},  reinforcement-learning~\cite{hahn2019, wang2019} and language parsing~\cite{mun2020} have been explored to improve video-description matching. However, a simple direct application of these methods on our VLN task does not perform well due to the visual representation difference:  consecutive frames with smooth transitions \textit{v.s.} sparse image sequences full of cutaways. And therefore, we design a new grounding method specifically for VLN tasks. 

\input{table/r2r}
\input{table/reverie}

\section{Method}
\noindent\textbf{Problem Formulation.}
In this work, we propose to minimize the SR and OSR gap of a given VLN model by formulating it as a trajectory grounding task. Given a natural language instruction $\mathcal{X} = [ x_1, x_2, \cdots,  x_L ] $ consisting of  $L$ words and a navigation trajectory $\mathcal{T}= [ \mathcal{V}_1, \mathcal{V}_2, \cdots \mathcal{V}_T ]$ consisting of $T$ viewpoints, trajectory grounding aims to find out the viewpoint that matches the specified destination in the instruction so as to reduce the gap between SR and OSR. Each viewpoint has a panorama observation represented by 36 discrete images $\mathcal{V}_t = \{\mathbf{v}_{t,i}\}_{i=1}^{36}$ (3 elevations $\times$ 12 headings) as in previous VLN works~\cite{r2r,reverie}. The trajectories are harvested via pre-trained VLN methods. Note that the destination  may not appear in the given trajectories. In this case, our model fails as well. But considering our aim is only to reduce the gap between SR and OSR in this work, this is acceptable.

~\newline\noindent\textbf{Method Overview.}
Figure~\ref{fig:architecture} illustrates the main architecture of the proposed model, which contains three transformer-based modules: Cross-Modality Elevation Transformer, Spatial-Temporal Transformer, and Target Selection Transformer. The model takes as  input an instruction-trajectory pair and utilizes a text encoder and a vision encoder to extract their own representations. Then, the representations are fed into the Cross-Modality Elevation Transformer that fuses information across elevations. Next, the Spatial-Temporal Transformer module exchanges information across discrete views and across temporal navigation steps. Finally, the Target Selection Transformer utilizes learnable queries to summarize the information of each viewpoint, and its outputs are used to predict the confidence of each viewpoint of being the destination. Below we give more details about each module.

\subsection{Text and Vision Encoder}

\noindent\textbf{Text Encoder.}
Given the natural language instruction $\mathcal{X}$, we utilize the linguistic embedding model BERT~\cite{bert} to exact features. Here, we use the BERT embedding of the first token (\textit{i.e.}, $\texttt{[CLS]}$ token) as the feature of instruction, denoted as $\mathbf{t}_{cls}$, which has been proven being able to represent comprehensive information of the whole instruction~\cite{hop,prevalent,duet}, and this can also significantly reduce computation and resource costs.

\noindent\textbf{Vision Encoder.}
A trajectory $\mathcal{T}$ is represented by a sequence of visual panorama observations at each viewpoint $\mathcal{T}  = [ \mathcal{V}_1, \mathcal{V}_2, \cdots \mathcal{V}_T ]$. Each panorama  $\mathcal{V}_t=\{\mathbf{v}_{t,i}\}_{i=1}^{36}$ consists of 36 discrete images (3 elevations and 12 headings in each elevation), each for one observing view (\textit{i.e.}, viewpoint). As in~\cite{hamt,duet}, we use ViT-B/16~\cite{vit} pre-trained on ImageNet~\cite{imagenet} to encode each image, obtaining its feature $\mathbf{o}_{t,{i}} \in \mathbb{R}^{1 \times d}$ and thus $\mathcal{O}_t \in \mathbb{R}^{36 \times d}$ for the viewpoint $\mathcal{V}_t$. Then, we add  navigation step encoding $\mathbf{E}_t^t \in \mathbb{R}^{36 \times d}$  and spatial sinusoidal positional encoding   $\mathbf{E}_t^s \in \mathbb{R}^{36 \times d}$ as~\cite{transformer, detr, tubedetr} onto each viewpoint representation:
\begin{equation}
{\mathcal{O}'_t} = \mathcal{O}_t + \mathbf{E}_t^t + \mathbf{E}_t^s.
\end{equation}
The trajectory-instruction pair representations $\langle {\mathbf {t}_{cls}, \mathcal{O}'_1}, \cdots, {\mathcal{O}'_T} \rangle$ are fed into the Cross-Modality Elevation Transformer.

\subsection{Cross-Modality Elevation Transformer}
\label{sec:spatialTemporal}
This module is designed to fuse information across elevations with the awareness of the instruction description. To this end, we first fuse vision-language information from the observing view level because the described destination is one of these views. Specifically, the feature $\mathbf{o}_{t,i}$ of each observing view is fused with the instruction embedding separately:
\begin{equation}
\mathbf{o}'_{t,i} = \textrm{FC}([\sigma(\mathbf{o}_{t,i} \mathbf{W}_o), \sigma(\mathbf{t}_{cls} \mathbf{W}_t)])
\end{equation}
\noindent where $\mathbf{W}_o$ and $\mathbf{W}_t$ are learnable parameter matrix, $\sigma$ denotes the activation function, and $[\cdot]$ denotes concatenation. 

Then, we further fuse the information across elevations. In each observing heading direction, there are three elevation views, as shown in Figure~\ref{fig:architecture}. Images of these elevations present significant overlap and thus  redundant information. To address this problem, we leverage a transformer module to fuse information based on attention mechanism. Another benefit of this module is that the number of vision tokens also sharply decreases to $1/3$ of the original. Specifically, observations at each heading angle can be denoted as $\mathbf{h}_{t,j} = \{\mathbf{o}'_{t,{j}}, \mathbf{o}'_{t,{j+12}}, \mathbf{o}'_{t,{j+24}}\}$, where $j \in [0,12)$. 
Then, we can reformulate $\mathcal{O}'_t$ in terms of $\mathbf{h}$: 
$\mathcal{O}'_t = \{\mathbf{h}_{t,j}\}_{j=0}^{11} \in \mathbb{R}^{12 \times 3 \times d}$. 
Next, we feed it to a stack of transformer layers with self-attention on each $\mathbf{h}_{t,j}$ to exchange information across elevations:
\begin{small}
\begin{gather}
\hat{\mathbf{h}}_{t,j}^E = \textrm{softmax}(\frac{{\mathbf{Q}_{t,j}^E}^\top\mathbf{K}_{t,j}^E}{\sqrt{d}})\mathbf{V}_{t,j}^E,\nonumber\\
\mathbf{Q}_{t,j}^E = {\mathbf{h}_{t,j}}\mathbf{W}_{e}^Q,\;\;
\mathbf{K}_{t,j}^E = {\mathbf{h}_{t,j}}\mathbf{W}_{e}^K, \;\;
\mathbf{V}_{t,j}^E = {\mathbf{h}_{t,j}}\mathbf{W}_{e}^V,
\end{gather}
\end{small}
~\newline\noindent where $\mathbf{W}_{e}^*$ is a learnable parameter matrix, and $d$ is the embedding dimension of  $\hat{\mathbf{h}}_{t,j}^E$.
Finally, we further aggregate  information across elevation via average pooling: $\widetilde{\mathbf{h}}_{t,j^E} = \textrm{AvgPool}(\hat{\mathbf{h}}_{t,j}^E)\in \mathbb{R}^{1 \times d}$ on elevation dimension. In this way, the instruction-informed representation at each trajectory viewpoint $\mathcal{O}'_t$ is converted to ${\mathcal{H}_t^E}=[\widetilde{\mathbf{h}}_{t,1}^E,\cdots,\widetilde{\mathbf{h}^E}_{t,12}]\in\mathbb{R}^{12 \times d}$. 

\subsection{Spatial-Temporal Transformer}
\label{sec:st-trm}
To make each visual token aware of the existence of other tokens in both spatial and temporal dimensions, we first perform spatial attention within each viewpoint  representation $\mathcal{H}_t^E\in\mathbb{R}^{12 \times d}$ across 12 headings:
\begin{equation}
\mathcal{H}^{H}_t = 
\textrm{softmax}(\frac{{({\mathcal{H}_t^E}\mathbf{W}^Q_h})^\top ({\mathcal{H}_t^E}\mathbf{W}^K_h)}{\sqrt{d}})
({\mathcal{H}_t^E}\mathbf{W}^V_h),
\end{equation}
where $\mathbf{W}_h^*$ are learnable parameters and $\mathcal{H}_t^H\in\mathbb{R}^{12 \times d}$. In this way, we obtain the updated representation for the whole trajectory $[\mathcal{H}_1^H,\cdots,\mathcal{H}_T^H] \in \mathbb{R}^{T\times 12 \times d}$.

Then, we expand the spatial attention from one viewpoint to all the viewpoints, attending to all $T \times 12$ observation representations:
\begin{equation}
\mathcal{H}^{ST} = 
\textrm{softmax}(\frac{{({\mathcal{H}^H}\mathbf{W}^Q_{st})}^\top ({\mathcal{H}^H}\mathbf{W}^K_{st})}{\sqrt{d}})
({\mathcal{H}^H}\mathbf{W}^V_{st}),
\end{equation}
~\newline\noindent where $\mathbf{W}^*_{st}$ are learnable parameters and $\mathcal{H}^{ST}\in\mathbb{R}^{T \times 12 \times d}$.
This helps the model to mine the long-range dependency in the whole trajectory. 

\subsection{Target Selection Transformer}
After cross-modality elevation and spatial-temporal transformers, the information of each viewpoint has been fully exchanged.
However, the representation dimension for a trajectory is still very high: $\mathbb{R}^{T \times 12 \times d}$, which leads to a large amount of computational cost for the end-to-end training of the whole model. First we use viewpoint query self-attention layer to make the $T$ input  queries attend to each other. To extract discriminative information, we leverage the query to adaptively summarize the information within each viewpoint representation ${\mathcal{H}_{t}^{ST}}$ via cross attention:
\begin{equation}
\hat{\mathbf{q}_{t}} = \textrm{softmax}(\frac{{({\mathbf{q}_{t}}\mathbf{W}_t^Q)}^\top ({\mathcal{H}_{t}^{ST}}\mathbf{W}_t^K)}{\sqrt{d}})
({\mathcal{H}_{t}^{ST}}\mathbf{W}_t^V)
\end{equation}
~\newline\noindent where  $\mathbf{W}^*_t$ are learnable parameters, and the dimension of the resulting $\hat{\mathbf{q}_{t}} $ is compressed to $ \mathbb{R}^{1\times d}$.
The final representation of the whole trajectory turns to $\hat{\mathcal{Q}} = [\hat{\mathbf{q}_1},\cdots,\hat{\mathbf{q}_T}] \in \mathbb{R}^{T\times d}$.
Next, we utilize a two-layer MLP followed by a sigmoid function to predict the  probability of each viewpoint being the described target location:
\begin{equation}
p_t = \mathrm{Sigmoid}(\mathrm{MLP}(\hat{\mathbf{q}_{t}})).
\end{equation}
Considering there might be more than one positive viewpoints in a trajectory,  we hope our model can discover all these viewpoints, so we view this as a step-wise two-class classification problem for each viewpoint. While during inference, the final prediction is determined by selecting the viewpoint with the maximum probability, which allows high prediction confidence.

\subsection{Loss Function}
\label{sec:loss}
We employ two loss functions for training: Focal Loss~\cite{focal} and Dice Loss~\cite{dice}. Focal Loss is a balanced version of cross-entropy loss that puts more emphasis on hard-to-classify examples. The focal loss function is defined as follows:
\begin{equation}
\mathcal{L}_{focal} = -\alpha_t(1-p_t)^\gamma \log(p_t),
\end{equation}
\noindent where $p_t$ is the predicted probability of the ground truth target viewpoint, $\alpha_t$ is a weighting factor that depends on the class frequency, and $\gamma$ is a hyper-parameter that controls how easy examples are downweighted. 

Dice Loss is a similarity-based loss function that measures the overlap between the prediction and ground truth labels:
\begin{equation}
\mathcal{L}_{dice} = 1 - \frac{2 \sum_{i=1}^{N} y_i p_t}{\sum_{i=1}^{N} y_i + \sum_{i=1}^{N} p_t},
\end{equation}
\noindent where $y_i$ is the binary ground truth label (1 for positive viewpoints and 0 for negative viewpoints).

The final loss is a weighted sum of Focal Loss and Dice Loss:
\begin{equation}
    \mathcal{L}_{total} = \lambda_{focal} \mathcal{L}_{focal} + \lambda_{dice} \mathcal{L}_{dice},
\end{equation}
\noindent where $\lambda_{focal}$ and $\lambda_{dice}$ are trade-off weights between Focal Loss and Dice Loss.

\input{table/ndh}

\section{Experiments}
We evaluate our proposed method on three downstream tasks: R2R~\cite{r2r}, REVERIE~\cite{reverie} and NDH~\cite{ndh} based on the Matterport3D~\cite{mattport} simulator. These tasks evaluate the agent from different perspectives and have different characteristics. 

\noindent\textbf{R2R benchmark.} The Room-to-Room (R2R) dataset~\cite{r2r} is a VLN task that involves navigating through photo-realistic indoor environments by following low-level natural language instructions, such as ``Turn left and walk across the hallway. Turn left again and walk across this hallway ...''. This task consists of 90 houses with 10,567 viewpoints and 7,189 shortest-path trajectories with 21,567 manually annotated instructions.

\noindent\textbf{REVERIE benchmark.} REVERIE dataset~\cite{reverie} is a VLN task that focuses on localizing a remote target object based on high-level human instructions, such as "Bring me the bench from the foyer". The target object is not visible at the starting location. The agent has to navigate to an appropriate location without detailed guidance.

\noindent\textbf{NDH benchmark.} NDH dataset~\cite{ndh} is a VLN task that evaluates an agent's ability to arrive at goal regions based on multi-turn question-answering dialogs. The task poses challenges due to the ambiguity and under-specification of starting instructions, as well as the long lengths of both instructions and paths.

\input{table/ab}

\subsection{Evaluation Metrics}
\label{sec:metrics}
We adopt widely-used metrics on each of these three VLN tasks. For the R2R task, \textbf{Trajectory Length} (TL) measures the average distance navigated by the agent. \textbf{Navigation Error} (NE) is used to measure the mean deviation between the agent's stop location and the target location in meters. The \textbf{Success Rate} (SR) is computed as the ratio of successfully completed tasks, where the agent stops within 3 meters to the target location, while \textbf{Oracle Success Rate} (OSR) considers a task as successful when at least one of its trajectory viewpoints is within 3 meters to the target location. \textbf{Success Rate weighted by Path Length} (SPL) evaluates the accuracy and efficiency of navigation simultaneously, which considers both the success rate and the length of the navigation path. For the REVERIE~\cite{reverie} task,   the same metrics as R2R are utilized to evaluate the navigation sub-task. REVERIE introduces additional metrics to evaluate the object grounding performance, \textbf{Remote Grounding Success Rate} (RGS) measures the proportion of tasks that successfully locate the target object, and \textbf{RGS weighted by Path Length} (RGSPL) which considers both grounding accuracy and the navigation path length. For the NDH~\cite{ndh} task, the primary evaluation metric in this study is \textbf{Goal Progress} (GP), which quantifies the distance (in meters) that the agent has progressed towards the target location. We also introduce the SR  of the R2R task  as a complementary measurement.

\subsection{Implementation Details}
We conducted all experiments on a single NVIDIA RTX 3090 GPU. The batch size is set to 48. The image features are exacted by ViT-B/16~\cite{vit} pretrained on ImageNet~\cite{imagenet}, and we adopt RoBERTa~\cite{roberta} as the text encoder. For each VLN benchmark, we trained our model using both generated data from the original data and HM3D-AutoVLN data~\cite{hm3d}. The Cross-Modality Elevation Transformer, Spatial-Temporal Transformer and Target Selection Transformer in our model use 2, 2 and 2 transformer layers, respectively. Our transformer has 8 heads, and feed-forward layers are with a hidden dimension of 768. We freeze the vision encoder and set the initial learning rates to $1\!\times\!10^{-5}$ for the language encoder and $1\!\times\!10^{-4}$ for the rest of the network. The text encoder has a linear schedule learning rate with warm-up while the rest of the network has a constant learning rate. We apply the AdamW~\cite{adamw} optimizer with weight-decay $1\!\times\!10^{-2}$ and use dropout probability of 0.1 in transformer layers and 0.5 in the prediction head. We also apply a dropout with a probability of 0.4 followed by a linear layer on image features. Lastly, we use an exponential moving average with a decay rate of 0.9998. For loss functions, we set hyper-parameters $\lambda_{focal}=1$, $\lambda_{dice}=0.1$, $\alpha_t=0.25$, and $\gamma=2$. We train our networks for 40000 iterations, and the final model is selected based on the best performance on the validation unseen split.

\noindent\textbf{Training Data Preparation.}
The ground-truth trajectories are not suitable for training our model because the target viewpoint is always the last one in each trajectory, which might lead to our model to learn this bias instead of the ability to match visual representation to textual ones. To overcome this problem, we construct new training data based on original trajectories. Specifically, for each original trajectory, we first find out all the viewpoints $Z$ that are located within 3 meters of the target location. These viewpoints are viewed as positive because if an agent stops at any of these viewpoints, the navigation is regarded as successful. Then, we  sample $1\!\sim\!2$ viewpoints (e.g., \{$\mathcal{A,B}$\}) from $Z$ as the positive viewpoints of a newly constructed trajectory. The reason we sample multiple positive viewpoints is that a VLN method may pass by the target location several times, such as before and after the target location. Next, we connect the original starting location with one of $\mathcal{A}$ and/or $\mathcal{B}$ using the shortest path. Then, from the other viewpoint (if there is any), we expand to another sub-path within 6 meters. By connecting these two sub-paths with $\mathcal{A}$ and/or $\mathcal{B}$, we obtain   a new path. In this way, we construct 396,866 trajectories for R2R,  331,533 trajectories for REVERIE, and 376,871 trajectories for NDH.

\input{table/add_data}

\subsection{Comparison to State-of-the-Art Methods}
\noindent\textbf{Results on R2R.} 
We assess the efficacy of the proposed method on top of four state-of-the-art methods: HOP~\cite{hop}, SIG\_HAMT~\cite{sig}, DUET~\cite{duet} and SIG\_DUET~\cite{sig}. The results are presented in Table~\ref{tab:r2r}. It shows that our method brings significant performance improvements for these four top-performing methods. Specifically, on validation unseen, our method improves the success rate with an absolute increase of 4\% for SIG\_DUET, 3\% for HOP and DUET, and 2\% for SIG\_HAMT. On the test split, our method yields 2\% absolute improvement for 3 out of the 4 top-performing methods, boosting the success rate to 74\%. In terms of SPL, we report two types of results: one is computed based on the path that adds the shortest return path from the original stop location to the viewpoint predicted by our method, and the other one is computed using the path cropped from the starting location to the viewpoint predicted by our method (indicated by superscript ${\spadesuit}$).
In the return setting, the SPL is slightly decreased (-1\%) due to the  increase in trajectory length caused by the return path. By contrast, in the crop setting, our method can bring 2\%$\sim$5\% absolute SPL improvement. All these results show the effectiveness of the proposed method.

\noindent\textbf{Results on REVERIE.}
In Table~\ref{tab:reverie}, we present results  on the REVERIE dataset with comparisons to several state-of-the-art methods, including RCM~\cite{rcm}, SMNA~\cite{selfmonitor}, FAST-Short~\cite{fast}, MATTN~\cite{reverie}, ORIST~\cite{orist}, \rvlnbert~\cite{recurrent}, and AirBERT~\cite{airbert}. REVERE evaluates both navigation and grounding performance. As our method is designed to improve navigation ability, the navigation success rate is the main metric for reference. As shown in Table~\ref{tab:reverie}, our method boosts the navigation success of all four top-performing baselines, with improvement ranging from 0.35\% to 2.33\%  across three splits. As our method only modifies pre-obtained navigation trajectories, it cannot change the object grounding results and thus there is no performance difference in terms of the RGS metric.

\input{table/loss}
\input{table/loss_hp}

\noindent\textbf{Results on NDH.}
The results of our baselines and our method are presented in Table~\ref{tab:ndh}. By applying our method to HAMT~\cite{hamt} and SIG\_HAMT~\cite{sig}, we achieved  2.65\% and 1.91\% improvements in success rate on the NDH validation unseen split, respectively. More improvements are observed on the validation seen split: 4.51\% and 3.77\%. Regarding the GP metric, our method also improve our baselines to the new state-of-the-art performance from 5.6 to 5.89.

\begin{figure*}[!t]
\centering
\includegraphics[width=0.93\linewidth]{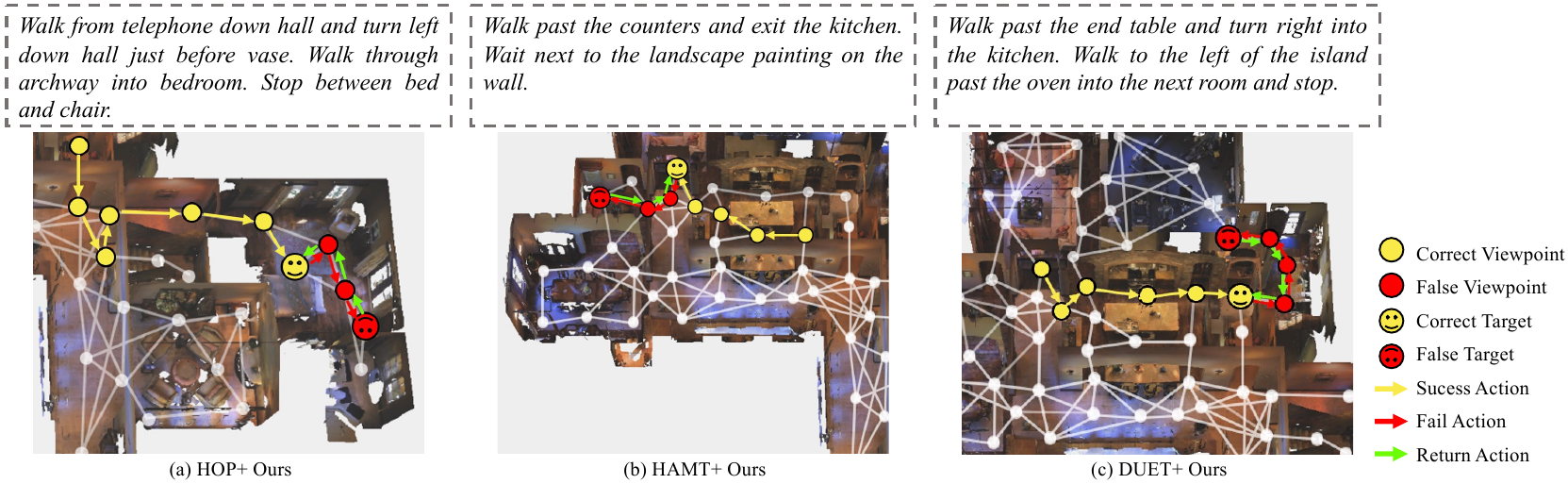}
\caption{Visualization of trajectories before and after applying our method on top of three method: HOP~\cite{hop}, HAMT~\cite{hamt}, and DUET~\cite{duet}. The yellow lines represent the successful part of the original path, while the red lines represent the incorrect part of navigation. The green lines show the return path from   an original stop location to the new stop location predicted by our methods.}
\label{fig:vis}
\end{figure*}

\subsection{Ablation Study}
In this section, we investigate the effectiveness of the three transformer-based modules of our method, additional training data,  loss functions, and hyperparameters of loss functions.

\noindent\textbf{The Effectiveness of Different Components.}
We conduct ablation experiments over the main components, cross-modality elevation transformer (E Trans.), spatial-temporal transformer (S-T Trans.), and target selection transformer (T Trans.), of our method on the R2R validation unseen split. The results are presented in Table~\ref{tab:ablation}. For comparison convenience, we also present the baseline result in \#0. In \#1, we present the results when applying all three modules onto the baseline. In the remaining rows, we report the results by removing each module one by one. It shows that, when applying all three modules, the best results are achieved and reduce the gap between SR and OSR from 7.96 to 4.56 (\#0 \textit{v.s.} \#1). When removing each module in turns, we notice that the elevator transformer leads to  the most performance drop (\#1 v.s. \#2), then the target selection transformer (\#1 \textit{v.s.} \#4), and last the spatial-temproal transformer (\#1 \textit{v.s.} \#3).

\noindent\textbf{The Effectiveness of Additional Training Data.} 
In addition to the original training data of each VLN task, we also use the automatically generated data in~\cite{hm3d} of 900 unlabelled raw HM3D environments. 
To verify its effectiveness, we conduct three ablation studies and the results are shown in Table~\ref{tab:add_data}. 
Specifically, using additional data resulted in a 2.17\% increase in Success Rate compared to the original training data (\#1 \textit{v.s.} \#3). 
We also evaluated zero-shot performance of the model trained only on additional data (\#2), which achieves a Success Rate of 45.52\% on the R2R validation unseen split.
Although the results suggest that the additional data from HM3D-VLN provides relevant information for the visual navigation task (\#2 \textit{v.s.} \#3), it is important to note that the model trained solely on the additional data achieves significantly lower performance than the model trained on only the original data. 

\noindent\textbf{Training Losses.} 
To effectively train our model, we test three loss functions: binary cross-entropy (BCE) loss, Focal loss~\cite{focal}, and Dice loss~\cite{dice}. Here we evaluate their own efficacy. The results are shown in Table~\ref{tab:loss}. As shown in the table, using focal loss with dice loss results in the best success rate, then BCE+Dice, Focal, and the BCE loss. Focal loss achieving better results than BCE loss indicates some training samples are important than others. The introduction of Dice loss boosts the performance over BCE and Focal loss. This indicates that considering multiple predictions together is better than handling them separately.

\noindent\textbf{Impact of Hyperparameters.}
We take another ablation study to investigate the impact of training loss hyperparameters (HPs) for our model. The ablation study involves varying the values of two HPs, $\lambda_{focal}$ and $\lambda_{dice}$, for the training loss function. Specifically, we experiment with four different combinations of hyper-parameters as shown in Table~\ref{tab:loss_hp}, ranging from both HPs having equal weight to one HP being given more weight than the other. The results  show that varying the hyper-parameters has a minor impact on the performance of our model.

\subsection{Qualitative Results}
In Figure~\ref{fig:vis},  we visualize the predicted trajectories   before and after applying our method on top of three methods: HOP~\cite{hop}, HAMT~\cite{hamt} and DUET~\cite{duet}. As shown in Figure~\ref{fig:vis}(a),  HOP initially passed the target location, which is a common problem in trajectory planning for VLN agents. Our method is able to find the correct viewpoints in the original trajectory and thus can correct wrong navigations as indicated by the dashed lines in Figure~\ref{fig:vis}. The similar phenomena can also be observed on HAMT~\cite{hamt} (Figure~\ref{fig:vis}(b)) and DUET~\cite{duet} (Figure~\ref{fig:vis}(c)). This shows the effectiveness of our method in correcting trajectories that deviate from desired path and enabling agents to reach their target locations with improved precision. 

\section{Conclusion}
In this paper, we address the long-ignored problem in Vision-and-Language Navigation of narrowing the gap between Success Rate (SR) and Oracle Success Rate (OSR) by formulating VLN as a trajectory grounding task. We propose to discover the target location from a trajectory given by off-the-shelf VLN models, rather than predicting actions directly as existing methods do. 
To achieve this, we design a novel multi-module transformer-based model to learn a compact discriminative trajectory viewpoint representation, which is then used to predict the confidence of being a target location. Our proposed approach is evaluated on four widely adopted datasets, R2R, REVERIE and NDH, with extensive evaluations demonstrating its effectiveness. Our method shows a new direction to solve the VLN tasks, which has been previously overlooked, and presents promising results for enhancing the performance of existing navigation agents.
\newpage
\balance
\input{paper.bbl}

\input{supp}
\end{document}

%% file: table/r2r.tex
\begin{table*}[!t]
\centering
\caption{Comparison with the state-of-the-art methods on R2R. \high{Blue}  and \textbf{Bold} denote the best and runner-up results, respectively. $\uparrow$ denotes the performance improvement after applying our method.  ${\spadesuit}$ denotes that we crop the original path from the starting location to the viewpoint predicted.}
\label{tab:r2r}
\resizebox{0.92\linewidth}{!}
{
\begin{tabular}{l|ccccc|ccccc|ccccc}
\toprule
    \multicolumn{1}{l|}{\multirow{2}{*}{Methods}} &\multicolumn{5}{c|}{R2R Val Seen} &\multicolumn{5}{c|}{R2R Val Unseen} &\multicolumn{5}{c}{R2R Test Unseen} \\ 
%\cmidrule(){2 - 16}
    ~ &TL&NE&SPL&SR&OSR  &TL&NE&SPL&SR&OSR  &TL&NE&SPL&SR&OSR\\ 
\midrule
    Human                     &-&-&-&-&-            &-&-&-&-&-            &11.85&1.61&76&86&90\\
\midrule
    Seq2Seq~\cite{r2r}        &11.33&6.01&-&39&53   &8.39&7.81&-&21&28    &8.13 &7.85&- &20&27\\
    SF~\cite{speakerfollower} &-    &3.36&- &66&74  &-    &6.62&- &36&45  &14.82&6.62&28&35&44\\
    RCM~\cite{rcm}            &10.65&3.53&- &67&75  &11.46&6.09&- &43&50  &11.97&6.12&38&43&50\\
    Regretful~\cite{regretful}&-    &3.23&63&69&-   &-    &5.32&41&50&-   &13.69&5.69&40&48&56\\
    FAST-short~\cite{fast}    &-    &-   &- &- &-   &21.17&4.97&43&56&-   &22.08&5.14&41&54&64\\
    EnvDrop~\cite{envdrop}    &11.00&3.99&59&62&-   &10.70&5.22&48&52&-   &11.66&5.23&47&51&59\\
    OAAM~\cite{oaam}          &10.20&-   &62&65&73  &9.95 &-   &50&54&61  &10.40&5.30&50&53&61\\
    EntityGraph~\cite{graph}  &10.13&3.47&65&67&-   &9.99 &4.73&53&57&-   &10.29&4.75&52&55&61\\
    NvEM~\cite{nvem}          &11.09&3.44&65&69&-   &11.83&4.27&55&60&-   &12.98&4.37&54&58&66\\
    ActiveVLN\cite{avig}      &19.70&3.20&52&70&-   &20.60&4.36&40&58&-   &21.60&4.33&41&60&- \\
    PRESS~\cite{press}        &10.57&4.39&55&58&-   &10.36&5.28&45&49&-   &10.77&5.49&45&49&- \\
    PREVALENT~\cite{prevalent}&10.32&3.67&65&69&-   &10.19&4.71&53&58&-   &10.51&5.30&51&54&61\\
    \rvlnbert~\cite{recurrent}  &11.13&2.90&68&72&79  &12.01&3.93&57&63&69  &12.35&4.09&57&63&70\\
    AirBERT~\cite{airbert}    &11.09&2.68&70&75&-   &11.78&4.01&56&62&-   &12.41&4.13&57&62&- \\
    SEvol~\cite{sevol}        &11.97&3.56&63&67&-   &12.26&3.99&57&62&-   &13.40&4.13&57&62&- \\
    ADAPT~\cite{adapt}        &10.97&2.54&72&76&-   &12.21&3.77&58&64&-   &12.99&3.79&59&65&- \\
    HAMT~\cite{hamt}          &11.15&2.51&72&76&82  &11.46&2.29&61&66&73  &12.27&3.93&60&65&72\\
\midrule
    HOP~\cite{hop} 
    &11.26&2.72&70&75&80  
    &12.27&3.80&57&64&71  
    &12.68&3.83&59&64&71\\
    {$+$}~Ours 
    &11.72&2.72&70&77($\uparrow$2)&80 
    &13.05&3.68&57&67($\uparrow$3)&71 
    &13.24&3.78&58&66($\uparrow$2)&71\\
    {$+$}~Ours$^{\spadesuit}$ 
    &9.95&2.72&\textbf{73}&77($\uparrow$2)&79 
    &9.69&3.68&62&67($\uparrow$3)&70 
    &10.23&3.78&61&66($\uparrow$2)&69\\
\midrule
    SIG\_HAMT~\cite{sig}    
    &11.68&2.80&70&73&79  
    &11.96&3.37&62&68&75  
    &12.83&3.81&60&65&72\\
    {$+$}~Ours 
    &12.24&2.80&69&76($\uparrow$3)&79 
    &12.62&3.36&62&70($\uparrow$2)&75 
    &13.42&3.73&59&66($\uparrow$1)&72\\
    {$+$}~Ours$^{\spadesuit}$ 
    &10.08&2.80&\textbf{73}&76($\uparrow$3)&78 
    &9.65&3.36&\textbf{66}&70($\uparrow$2)&73 
    &10.09&3.73&\textbf{62}&66($\uparrow$1)&70\\
\midrule
    DUET~\cite{duet}          
    &12.32&2.28&\textbf{73}&\textbf{79}&{86}          
    &13.94&3.31&60&72&{81}     
    &14.73&3.65&59&69&76\\
    {$+$}~Ours 
    &12.83&2.33&72&\high{82}($\uparrow$3)&{86}                      
    &14.65&3.27&60&\textbf{75}($\uparrow$3)&{81}            
    &15.33&3.62&58&71$(\uparrow$2)&76\\
    {$+$}~Ours$^{\spadesuit}$ 
    &11.24&2.33&\high{77}&\high{82}($\uparrow$2)&{85}       
    &11.95&3.27&65&\textbf{75}($\uparrow$3)&{79}      
    &12.98&3.62&61&71$(\uparrow$2)&75\\
\midrule
    SIG\_DUET~\cite{sig}    
    &13.96&2.73&67&75&83
    &14.31&3.13&62&72&{81} 
    &15.36&3.37&60&\textbf{72}&{80}\\
    {$+$}~Ours 
    &14.68&2.73&66&\textbf{79}($\uparrow$4)&{83} 
    &15.09&3.10&61&\high{76}($\uparrow$4)&{81}    
    &16.02&3.31&59&\high{74}($\uparrow$2)&{80}\\
    {$+$}~Ours$^{\spadesuit}$ 
    &11.81&2.73&72&\textbf{79}($\uparrow$4)&81
    &11.90&3.10&\high{67}&\high{76}($\uparrow$4)&{79}    
    &12.88&3.31&\high{64}&\high{74}($\uparrow$2)&{78} \\
\bottomrule
\end{tabular}
}
\end{table*}

%% file: table/reverie.tex
\begin{table*}[!t]
    \centering
    \caption{Comparison with the state-of-the-art methods on REVERIE. \high{Blue}  and \textbf{Bold} denote the best and runner-up results, respectively. $\uparrow$ denotes the performance improvement after applying our method. ${\spadesuit}$ denotes that we crop the original path from the starting location to the viewpoint predicted.}
    \label{tab:reverie}
    \resizebox{1.0\linewidth}{!}
    {
        \begin{tabular}{l|ccc|cc|ccc|cc|ccc|cc}
            \toprule
                \multicolumn{1}{l|}{\multirow{3}{*}{Methods}} &\multicolumn{5}{c|}{REVERIE Val Seen} &\multicolumn{5}{c|}{REVERIE Val Unseen} &\multicolumn{5}{c}{REVERIE Test Unseen} \\
                ~ &\multicolumn{3}{c}{Navigation}&\multicolumn{1}{c}{\multirow{2}{*}{RGS}}&\multicolumn{1}{c|}{\multirow{2}{*}{RGSPL}} &\multicolumn{3}{c}{Navigation}&\multicolumn{1}{c}{\multirow{2}{*}{RGS}}&\multicolumn{1}{c|}{\multirow{2}{*}{RGSPL}} &\multicolumn{3}{c}{Navigation}&\multicolumn{1}{c}{\multirow{2}{*}{RGS}}&\multicolumn{1}{c}{\multirow{2}{*}{RGSPL}} \\
                ~&\multicolumn{1}{c}{SPL}&\multicolumn{1}{c}{SR}&\multicolumn{1}{c}{OSR}&&           &\multicolumn{1}{c}{SPL}&\multicolumn{1}{c}{SR}&\multicolumn{1}{c}{OSR}&& &\multicolumn{1}{c}{SPL}&\multicolumn{1}{c}{SR}&\multicolumn{1}{c}{OSR}&& \\
        \midrule
            Human &-&-&-&-&-  &-&-&-&-&-  &53.66&81.51&86.83 &77.84&51.44 \\
        \midrule
            RCM \cite{rcm}           &21.82&23.33&29.44 &16.23&15.36  &6.97 &9.29 &14.23 &4.89 &3.89   &6.67 &7.84 &11.68 &3.67 &3.14 \\
            SMNA \cite{selfmonitor}  &39.61&41.25&43.29 &30.07&28.98  &6.44 &8.15 &11.28 &4.54 &3.61   &4.53 &5.80 &8.39  &3.10 &2.39 \\
            FAST-Short \cite{fast}   &40.18&45.12&49.68 &31.41&28.11  &6.17 &10.08&20.48 &6.24 &3.97   &8.74 &14.18&23.36 &7.07 &4.52 \\
            MATTN \cite{reverie}&45.50&50.53&55.17 &31.97&29.66  &7.19 &14.40&28.20 &7.84 &4.67   &11.61&19.88&30.63 &11.28&6.08 \\
            ORIST~\cite{orist}       &42.21&45.19&49.12 &29.87&27.77  &15.14&16.84&25.02 &8.52 &7.58   &18.97&22.19&29.20 &10.68&9.28 \\
            \rvlnbert~\cite{recurrent} &47.96&51.79&53.90 &38.23&35.61  &24.90&30.67&35.02 &18.77&15.27  &23.99&29.61&32.91 &16.50&13.51\\
            AirBERT~\cite{airbert}   &42.34&47.01&48.98 &32.75&30.01  &21.88&27.89&34.51 &18.23&14.18  &23.61&30.28&34.20 &16.83&13.28\\

        \midrule
            HOP~\cite{hop}
            &47.19&53.76&54.88 &38.65&33.85  
            &26.11&31.78&36.24 &18.86&15.73  
            &24.34&30.17&33.06 &17.69&14.34\\
            {$+$}~Ours               
            &46.43&54.11($\uparrow$0.35)&54.88&38.65&33.56  
            &25.79&32.83($\uparrow$1.05)&36.24&18.86&15.71  
            &24.28&31.29($\uparrow$1.12)&33.06&17.69&14.12  \\
            {$+$}~Ours$^{\spadesuit}$
            &50.61&54.11($\uparrow$0.35)&27.64&38.65&26.23  
            &54.63&32.83($\uparrow$1.05)&36.09&18.86&32.98  
            &35.12&31.29($\uparrow$1.12)&36.09&17.69&14.51  \\
        \midrule
            HAMT~\cite{hamt}          
            &40.19&43.29&47.65&27.20&25.18  
            &30.16&32.95&36.84&18.92&17.28  
            &26.67&30.40&33.41&14.88&13.08  \\
            {$+$}~Ours                
            &39.73&44.60($\uparrow$1.31)&47.65&27.20&24.90  
            &29.56&34.17($\uparrow$1.22)&36.84&18.92&16.93  
            &26.61&31.19($\uparrow$0.79)&33.41&14.88&12.82  \\
            {$+$}~Ours$^{\spadesuit}$ 
            &42.37&44.60($\uparrow$1.31)&47.40&27.20&25.37  
            &31.58&34.17($\uparrow$1.22)&36.68&18.92&17.56  
            &27.92&31.19($\uparrow$0.79)&33.28&14.88&13.47  \\
        \midrule
            DUET~\cite{duet}          
            &\textbf{63.94}&\textbf{71.75}&73.86&57.41&51.14  
            &33.73&46.98&51.07&32.15&23.03  
            &36.06&52.51&56.91&31.88&22.06  \\
            {$+$}~Ours                
            &63.21&\high{72.48}($\uparrow$0.73)&73.86&57.41&50.92  
            &33.57&48.31($\uparrow$1.33)&51.07&32.15&22.84  
            &35.30&54.08($\uparrow$1.57)&56.91&31.88&21.84  \\
            {$+$}~Ours$^{\spadesuit}$ 
            &\high{66.80}&\high{72.48}($\uparrow$0.73)&73.46&57.41&53.18  
            &34.96&48.31($\uparrow$1.33)&50.93&32.15&23.33  
            &38.19&54.08($\uparrow$1.57)&56.56&31.88&22.18  \\
        \midrule
           HM3D~\cite{hm3d}   
           &55.70&65.00&66.76&48.42&41.67  
           &\textbf{40.84}&\textbf{55.89}&62.14&36.58&26.75  
           &\textbf{38.88}&\textbf{55.17}&62.30&32.23&22.68  \\
           {$+$}~Ours                
           &55.60&65.50($\uparrow$0.50)&66.76&48.42&41.48  
           &40.61&\high{58.22}($\uparrow$2.33)&62.14&36.58&26.52  
           &38.72&\high{56.78}($\uparrow$1.61)&62.30&32.23&22.53  \\
           {$+$}~Ours$^{\spadesuit}$ 
           &58.22&65.50($\uparrow$0.50)&66.54&48.42&43.69  
           &\high{43.04}&\high{58.22}($\uparrow$2.33)&61.73&36.58&26.98  
           &\high{41.76}&\high{56.78}($\uparrow$1.61)&61.82&32.23&23.03  \\
        \bottomrule
        \end{tabular}
    }
\end{table*}

%% file: table/ndh.tex
\begin{table}[!t]
\centering
\caption{Comparison with the state-of-the-art methods on NDH validation seen and unseen splits. \high{Blue}  and \textbf{Bold} denote the best and runner-up results, respectively.  $\uparrow$ denotes the performance improvement after applying our method.}
\label{tab:ndh}
\resizebox{1\linewidth}{!}{
\begin{tabular}{l|ccc|ccc}
\toprule
    \multicolumn{1}{l|}{\multirow{2}{*}{Methods}} &\multicolumn{3}{|c}{NDH Val Seen} &\multicolumn{3}{|c}{NDH Val Unseen} \\ 
    ~ &GP&SR&OSR &GP&SR&OSR \\ 
\midrule
    PREVALENT~\cite{prevalent} &-&-&-  &3.15&-&- \\
    HOP~\cite{hop}           &-&-&-  &5.13&-&- \\
\midrule
    HAMT~\cite{hamt}        &6.90&20.68&31.15    &5.09&{16.87}&27.89   \\
    + Ours                  &7.28($\uparrow$0.38) &\textbf{25.19}($\uparrow$4.51)&31.15 &5.52($\uparrow$0.43)&\high{19.52}($\uparrow$2.65)&27.89 \\
\midrule
    SIG\_HAMT~\cite{sig}&\textbf{8.13}&23.56&33.77                       &\textbf{5.60}&15.33&28.45 \\
    + Ours              &\high{8.51}($\uparrow$0.38) &\high{27.33}($\uparrow$3.77) &33.77  &\high{5.89}($\uparrow$0.29)&\textbf{17.24}($\uparrow$1.91)&28.45 \\
\bottomrule
\end{tabular}}

\end{table}

%% file: table/ab.tex
\begin{table}[!t]
\centering
\caption{Ablation study on R2R validation unseen split using HOP~\cite{hop} as our baseline.
E Trans. denotes the cross-modality elevation transformer, S-T Trans. means the spatial-temporal transformer, and T Trans. represents target selection transformer.}
\label{tab:ablation}
\resizebox{0.90\linewidth}{!}{
\begin{tabular}{cc ccc |cc|c}
\toprule
 \multicolumn{1}{c}{\multirow{2}{*}{Model}}& &\multicolumn{3}{c|}{Component} & \multicolumn{3}{c}{R2R Val Unseen}\\
&\#  &E Trans. &S-T Trans. &T Trans. &SR &OSR&Gap \\ 
\midrule
HOP~\cite{hop} & 0  & & & & 63.52  &71.48 &7.96 \\
\midrule
\multicolumn{1}{c}{\multirow{4}{*}{{$+$}~Ours}}

&1  &\ding{51} &\ding{51} &\ding{51} &66.92 & 71.48 &4.56 \\
&2  &\ding{55} & & & 65.79  & 71.48 &5.69\\
&3  & &\ding{55} & & 64.82  & 71.48 &6.66\\
&4  & & &\ding{55} & 65.33  & 71.48 &6.15\\

\bottomrule
\end{tabular}}
\end{table}

%% file: table/add_data.tex
\begin{table}[t]
%\small
\centering
\caption{The performance of HOP~\cite{hop} with our model on R2R validation unseen split obtained via models trained with different data.}
\label{tab:add_data}
\resizebox{0.74\linewidth}{!}{
\begin{tabular}{c cc |cc|c}
\toprule
\multicolumn{1}{c}{\multirow{2}{*}{Model}} &\multicolumn{2}{c|}{Training Data} &\multicolumn{3}{c}{R2R Val Unseen}\\ 
&R2R &HM3D~\cite{hm3d}  &SR &OSR&Gap \\ 
\midrule
HOP~\cite{hop} & &       & 63.52 & 71.43 &7.96 \\
\midrule
1 &\ding{51} &           & 64.75 & 71.48 &6.73\\
2 &           &\ding{51} & 45.52 & 71.48 &25.96\\
3 &\ding{51} &\ding{51} &{66.92} & 71.48 &{4.56}\\
\bottomrule
\end{tabular}
}
\end{table}

%% file: table/loss.tex
\begin{table}[t]
\centering
\caption{The performance of HOP~\cite{hop} with our model on R2R validation unseen split with different training losses.}
\label{tab:loss}
\resizebox{0.67\linewidth}{!}{
\begin{tabular}{c l| cc|c}
\toprule
 \multicolumn{1}{c}{\multirow{2}{*}{Model}} &\multicolumn{1}{c|}{\multirow{2}{*}{Trainning Loss}} & \multicolumn{3}{c}{R2R Val Unseen}\\
 & &SR &OSR&Gap \\ 

\midrule

1 &BCE & 63.62 & 71.48 &7.86\\
2 &Focal & 65.05 & 71.48 &6.43\\
3 &BCE~{$+$}~Dice & 65.43 & 71.48 &6.05\\

4 &Focal~{$+$}~Dice &{66.92} & 71.48 &{4.56}\\

\bottomrule
\end{tabular}
}

\end{table}

%% file: table/loss_hp.tex
\begin{table}[t]
%\small
\centering
\caption{The performance of HOP~\cite{hop} with our model on R2R validation unseen split with different training loss hyper-parameters (HP).}
\label{tab:loss_hp}
\resizebox{0.67\linewidth}{!}{
\begin{tabular}{c cc |cc|c}
\toprule
 \multicolumn{1}{c}{\multirow{2}{*}{Model}}&\multicolumn{2}{c|}{Training Loss HP} & \multicolumn{3}{c}{R2R Val Unseen}\\
 & $\lambda_{focal}$ & $\lambda_{dice}$  &SR &OSR&Gap \\ 

\midrule

1 &1.0  &1.0   & 65.42 & 71.48 &6.06\\
2 &1.0  &0.5 & 65.53 & 71.48 &5.95\\
3 &1.0  &0.2 & 66.31 & 71.48 &5.17\\
4 &1.0  &0.1 &{66.92} & 71.48 &{4.56}\\

\bottomrule
\end{tabular}
}

\end{table}

%% file: paper.bbl

%% file: supp.tex
% \documentclass[sigconf,nonacm]{acmart}

% \usepackage{hyperref}
% \usepackage{multirow}
% \usepackage{enumitem}

% \usepackage[T1]{fontenc}
% \let\widebar\relax
% \let\bigtimes\relax
% \usepackage{mathabx}
% \def\CircleArrowright{\ensuremath{\rotatebox[origin=c]{310}{$\circlearrowright$}}}
% \newcommand{\rvlnbert}{VLN$\protect\CircleArrowright$BERT}

% \usepackage{pifont}
% \newcommand{\cmark}{\textcolor{cyan}{\ding{51}}}
% \newcommand{\xmark}{\textcolor{red}{\ding{55}}}
% \newcommand{\high}[1]{{\textbf{\color[RGB]{0,0,255}#1}}}

% \AtBeginDocument{\providecommand\BibTeX{{Bib\TeX}}}

% \begin{document}

% \title{Supplementary for: \textit{"Mind the Gap: Improving Success Rate of Vision-and-Language Navigation by Revisiting Oracle Success Route"}}

% \maketitle
\appendix
\section*{Appendix}

The appendix is organized as follows:
\begin{itemize}
    \item In Section~\ref{additional_results}, we provide additional results of our method on RxR validation splits. 
    
    \item In Section~\ref{additional_ablations}. we present more ablation studies of our MG-VLN model. 
    
    \item In Section~\ref{additional_qualitative}, we present several examples of our constructed new trajectories.
\end{itemize}

\section{Additional Results}
\label{additional_results}

\noindent\textbf{RxR benchmark.} 
The Room-Across-Room (RxR)~\cite{rxr} dataset is a large multilingual VLN dataset that comprises a vast collection of elaborate instructions and trajectories, including instructions in three different languages: English, Hindi, and Telugu. 
The dataset emphasizes the role of language in VLN by addressing biases in paths and describing more visible entities than R2R~\cite{r2r}. 
We construct 427,347 trajectories for RxR using the same method mentioned in the main paper. 

~\newline\noindent\textbf{Evaluation Metrics.} 
For the RxR~\cite{rxr}, \textbf{Normalized Dynamic Time Warping} (nDTW)~\cite{ndtw} and \textbf{Success Rate weighted by Dynamic Time Warping} (sDTW) are introduced in addition to the SR and SPL metrics mentioned in R2R~\cite{r2r}, to assess path fidelity.

~\newline\noindent\textbf{Results on RxR.} The results are presented in Table~\ref{tab:rxr}. 
By applying our method to HAMT~\cite{hamt}, we achieved 3.2\% and improvements in success rate on the RxR validation unseen split, respectively.
More improvements are observed on the validation seen split: 3.8\%. 
This demonstrates that our method can improve the navigation success capability of the VLN agent.

\input{table/rxr}

\begin{figure}[ht]
    \centering 
    \includegraphics[width=1\linewidth]{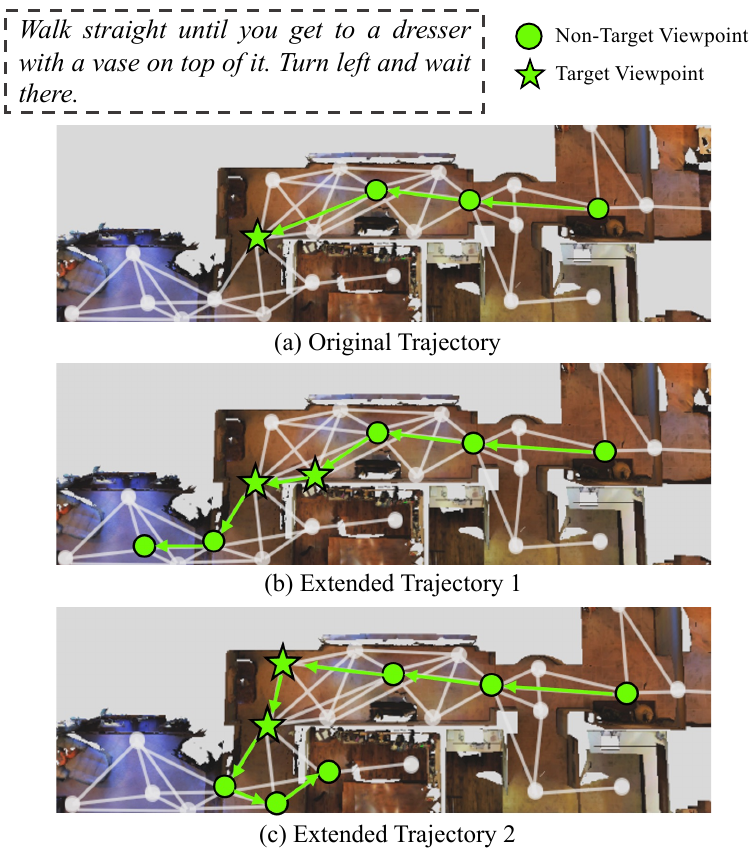}
	
    \caption{Examples of our constructed trajectories based on the ground-truth trajectory.
    }
    \label{fig:example}
\end{figure}

\section{Additional Ablations}
\label{additional_ablations}
\input{table/id}
\noindent\textbf{Accuracy in identifying the correct destination.}
The results of the accuracy in identifying the correct destination of our model with HOP~\cite{hop} on R2R Val Unseen split are presented in Table~\ref{tab:id}. \textit{"Keeping SR"} denotes the percentage of episodes that are still successful after our correction. \textit{"Correcting Gap"} denotes the percentage of episodes that are previously failed but become successful after our correction.

\input{table/cls}
\noindent\textbf{Can Text [CLS] Encode Destination Information?}
To investigate whether the Text $\texttt{[CLS]}$ token (Section 3.1 Text Encoder. in the main paper) can encode the destination of an instruction after fine-tuning, we conducted an ablation study as presented in Table~\ref{tab:cls}. 
Specifically, we compare the model performance between feeding the full instruction and the sub-instruction that only contains  information related to the target location. 
The results show that the performances are very close with a small gap of 0.6\%. 
The results suggest that the $\texttt{[CLS]}$ token can learn the  destination information of an instruction.

\section{Constructed Trajectories}
\label{additional_qualitative}
We provide examples of constructed new trajectories based on ground truth trajectory. 
Figure~\ref{fig:example} (a) illustrates an original ground truth trajectory, where the target viewpoint is represented by a pentagram. 
Figure~\ref{fig:example} (b) and (c) present examples of new trajectories that are constructed based on the original trajectory.

% \balance
% \bibliographystyle{ACM-Reference-Format}
% \bibliography{ref_list}
% \end{document}
% \endinput

%% file: table/rxr.tex
\begin{table}[ht]
\centering
\caption{Comparison with the state-of-the-art methods on Room-Across-Room (RxR) validation seen and unseen splits. $\uparrow$ denotes the performance improvement after applying our method. ${\spadesuit}$ denotes that we crop the original path from the starting location to the viewpoint predicted.}
\label{tab:rxr}
\resizebox{1\linewidth}{!}{
\begin{tabular}{l|ccccc|ccccc}
\toprule
    \multicolumn{1}{l|}{\multirow{2}{*}{Methods}} &\multicolumn{5}{|c}{RxR Val Seen} &\multicolumn{5}{|c}{RxR Val Unseen} \\ 
    ~ &SR&OSR&SPL&nDTW&SDTW &SR&OSR&SPL&nDTW&SDTW \\ 
\midrule
    Baseline ~\cite{rxr} &25.2&-&-&42.2&20.7  &22.8&-&-&38.9&18.2 \\
\midrule
    HAMT~\cite{hamt}          &59.4&66.5&58.9&65.3&50.9  &56.6&64.4&56.0&63.1&48.3   \\
    {$+$}~Ours                &63.2($\uparrow$3.8)&66.5&58.1&66.9&52.7  &59.8($\uparrow$3.2)&64.4&55.2&63.9&49.2 \\
    {$+$}~Ours$^{\spadesuit}$ &63.2($\uparrow$3.8)&66.1&60.2&67.2&53.0  &59.8($\uparrow$3.2)&63.5&56.7&64.4&49.6 \\
\bottomrule
\end{tabular}}
\end{table}

%% file: table/id.tex
\begin{table*}[!t]
\centering
\caption{The accuracy in identifying the correct destination of our model with HOP~\cite{hop} on R2R validation unseen split.}
\label{tab:id}
\resizebox{0.8\linewidth}{!}{
\begin{tabular}{cl|llll}
\toprule
\multicolumn{2}{l|}{R2R Val Unseen} & \multicolumn{1}{c}{HOP~\cite{hop}} & \multicolumn{1}{c}{SIG\_HAMT~\cite{sig}} & \multicolumn{1}{c}{DUET~\cite{duet}} & \multicolumn{1}{c}{SIG\_DUET~\cite{sig}}   \\ 
\midrule
0 &Keeping SR     &98.12\% (62.32)    &98.56\% (67.09) &98.75\% (70.63)	&98.35\% (71.22) \\
1 &Correcting Gap &67.08\% (4.60)     &58.52\% (3.36) &51.67\% (3.96)	&62.15\% (4.68) \\
2 &Final SR       &66.92 (62.32+4.6)  &70.46 (67.09+3.36)	&74.58(70.63+3.96)	&75.9 (71.22+4.68) \\

\bottomrule
\end{tabular}}

\end{table*}

%% file: table/cls.tex
\begin{table}[ht]
\centering
\caption{Ablation study of the [CLS] token on the R2R validation unseen split with different instruction inputs.}
\label{tab:cls}
\resizebox{0.75\linewidth}{!}{
\begin{tabular}{c l| cc|c}
\toprule
 \multicolumn{1}{c}{\multirow{2}{*}{Methods}} &\multicolumn{1}{c|}{\multirow{2}{*}{Input Text}} & \multicolumn{3}{c}{R2R Val Unseen}\\
 & &SR &OSR&Gap \\ 
\midrule
HOP~\cite{hop} & & 63.52 & 71.43  & 7.96\\
\midrule

+ Ours (1) &Target Instruction & 66.39 & 71.48 &5.09\\

+ Ours (2) &Whole Instruction & 66.92 & 71.48 &4.56\\

\bottomrule
\end{tabular}}

\end{table}